# Automatic Business Process Structure Discovery using Ordered Neurons LSTM: A Preliminary Study

Xue Han[1], Lianxue Hu[1], Yabin Dang[1], Shivali Agarwal[2], Lijun Mei[1], Shaochun Li[1], Xin Zhou[1]

1. IBM Research China {bjhanxue, hulxue, dangyb, meilijun, lishaoc, zhouxin}@cn.ibm.com
2. IBM Research India shivaaga@in.ibm.com

**Abstract**

Automatic process discovery from textual process documentations is highly desirable to reduce time and cost of Business Process Management (BPM) implementation in organizations. However, existing automatic process discovery approaches mainly focus on identifying activities out of the documentations. Deriving the structural relationships between activities, which is important in the whole process discovery scope, is still a challenge. In fact, a business process has latent semantic hierarchical structure which defines different levels of detail to reflect the complex business logic. Recent findings in neural machine learning area show that the meaningful linguistic structure can be induced by joint language modeling and structure learning. Inspired by these findings, we propose to retrieve the latent hierarchical structure present in the textual business process documents by building a neural network that leverages a novel recurrent architecture, Ordered Neurons LSTM (ON-LSTM), with process-level language model objective. We tested the proposed approach on data set of Process Description Documents (PDD) from our practical Robotic Process Automation (RPA) projects. Preliminary experiments showed promising results.

## 1. Introduction

With potential to reduce costs, improve productivity and achieve higher levels of quality, Business Process Management (BPM) has been attracting more and more attention recently (Mendling and Ingo 2018). A BPM life cycle starts with the process discovery phase which focuses on producing detailed descriptions of a business process as it currently exists in a structured form like the "as-is" process model. On base of this "as-is" model, methods and techniques are developed to analyze the current weaknesses and redesign the "to-be" process model until finally lead to improvements on the business process (M. J. Dumas Marlon 2013).

Although key to a successful BPM project, the discovery of "as-is" business process model is very labor intensive and time-consuming. It requires a thorough analysis of current process related knowledge and information, 85% of which are estimated to be available in an unstructured or less structured textual form such as memos, manuals, requirements documents, design documents, etc. (A. Ghose and Chueng 2007). For convenience, we use documents to represent all these forms in the following sections. Therefore, it is highly desirable to make process discovery automatically.

To achieve this expectation, researches have proposed automatic process discovery methods aiming at helping analysts to create better models from large number of documents in less time (H.der Aa and et al. 2018). These methods (H.Leopold and H.Reijers 2018; Epure Elena Viorica; Martin-Rodilla 2015; Delicado and Josep 2017; Sintoris and Vergidis 2017) mainly firstly use NLP (Natural Language Processing) pipelines to make syntactic analysis of natural language text at word or sentence level to identify activities out of documents. Then leverage the labeled process document features (such as typical patterns or documents layout) to describe the structural relationships between activities. However, for large scale organizations with hundreds or thousands of different processes, the effort required to identify and label such features is considerable. Therefore, deriving the business process structure, which is important in the whole process discovery scope, is still a challenge.

In fact, a business process has latent hierarchical structure which defines different levels of detail to show the processes of a company and reflect the complex business logic. Such latent hierarchical structure is the reason why formal and graphical process notations, such as Event-driven Process Chains (EPCs), Business Process Model and Notation (BPMN) etc., are developed for modeling unambiguous representations of business processes in addition to documents (M.Dumas and HA.Reijers 2013).

To be specific, many BPM supporting platforms are designed to reflect such latent hierarchical structure. Take the recently emerging Robotic Process Automation (RPA) as an example. RPA solution can be perceived in layers (as illustrated in Figure1(a)), starting with the process layer which can also call another as a child, down to the object layer which call APIs provided by target applications (such as SAP). An example of Process Definition Document (PDD) which is used to describe the 'as is' manual process (following the template of BluePrism[1] is illustrated as in Figure 1(b). According to Figure 1(b), the textual description contains layers corresponding to Figure 1(a). Compared with graphical process models, documents usually include more sentences to further explain the activities and help the reader understand the process clearly. Examples of sentences are like sentences 1.4 and 1.5 in Figure1(b). The latent hierarchy structure of Figure1(b) is represented as Figure 1(c), in which the root is a parent process, and the child sentences explain their parent in detail.

---

1. https://www.blueprism.com/support/

In this paper, we are aiming at discovering such latent process structure automatically without extra human labeled knowledge. Recent findings in neural machine learning area show that the meaningful linguistic structure can be induced by joint language modeling and structure learning. (Shen et al. 2019; Dongyeop Kang and Hovy 2019; Hewitt and Manning 2019; Masaru Isonuma and Sakata 2019; Jie Hao and et al. 2019). Inspired by these findings, we propose an approach to learn the latent process structure present in documents by leveraging the business process language modeling without any further expert knowledge. We firstly extend the sentence-level language model to process-level. Then we train a neural network on base of an advanced variant of RNNs-Ordered Neurons LSTM (ON-LSMT) (Shen et al. 2019) using the process-level language modeling objective. Finally, we retrieve the latent hierarchical structure present in the textual business process documents from the trained model. We tested the proposed approach on data set of PDD (Process Description Documents) from our practical RPA projects which are designed following BluePrism guidelines. Preliminary experiments showed promising results.

The paper is structured as follows. Section 2 introduces related work. Section 3 illustrates the framework of our proposed approach and introduces details of each component in the framework. Section 4 presents our evaluation results based on RPA documents. Section 5 discusses the future work.

## 2. Related Work

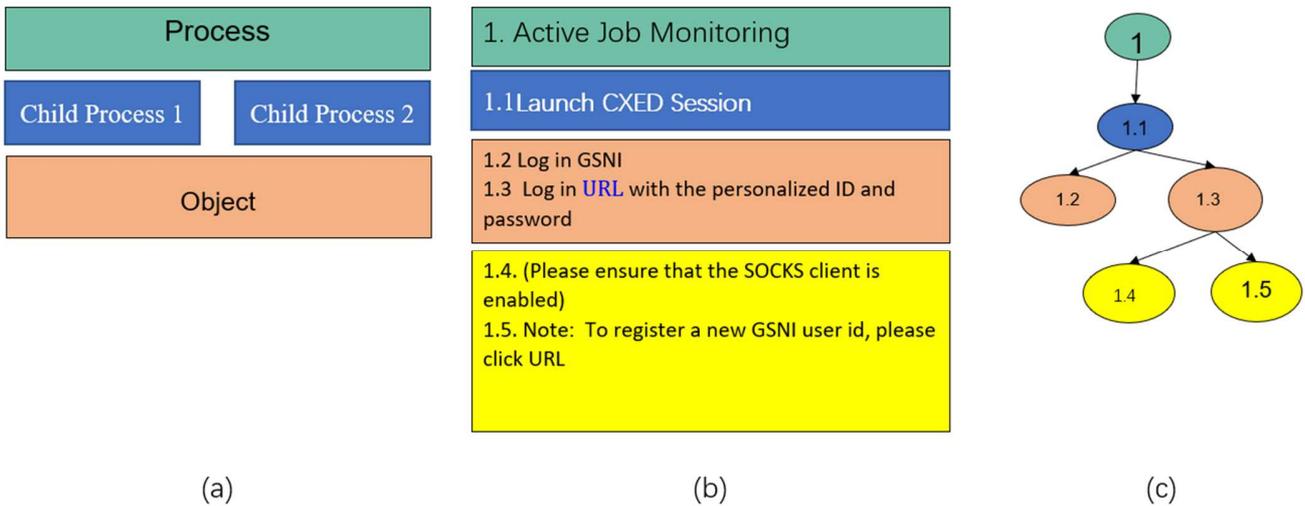

Figure 1: (a) RPA solution design in layers; (b) An example of Process Definition Document which is used to describe the 'as is' manual process; (c)The latent hierarchy structure

Existing methods mainly leverage the labeled process document features (such as typical patterns or documents layout) to describe structural relations. (Epure Elena Viorica; Martin-Rodilla 2015; Delicado and Josep 2017; Sintoris and Vergidis 2017) proposed similar rule-based systems to mine the relationships between activities. These rules can either be a single word like "if", "then", "meanwhile", or a short phrase like "in the meantime" or "in parallel", which are handcrafted on base of human labeled knowledge. It requires a lot of effort to identify and label such patterns in large organizations (H.der Aa and A.Henrik 2017). Also handcraft patterns can't understand the complex semantic information in the documents. (M.Mohammad and et al. 2018) proposed to construct structural formatted data, by extracting document layout features and then analyzing the changes in the layout features. However, according to practical experience, it is hard to depend on the authors of documents to describe a process strictly following the layout rules. In addition, documents representation is not standardized and displays a great deal of variability, making the analysis of the layout features even more complicate.

## 3. Approach

We propose to extract the latent hierarchical structure out of documents conditioned by building a neural network that leverages ON-LSTM (Shen et al. 2019) network with process-level language model objective. The overall archi-tecture of the proposed approach is as Figure 2. Our approach contains a process encoder, a 3-layer ON-LSTM network and a structure retrieval component. Given a paragraph of process description P D = (S$_1$, S$_2$, …, S$_L$) containing L sentences, the sentence sequences are firstly encoded in the process encoder. Then ON-LSTM network takes the process encoded vectors as input to train a process-level language model. Finally, the parameters and hidden vectors of the trained model could be conditioned to retrieve the latent process structure described in documents without any extra human labeled knowledge.

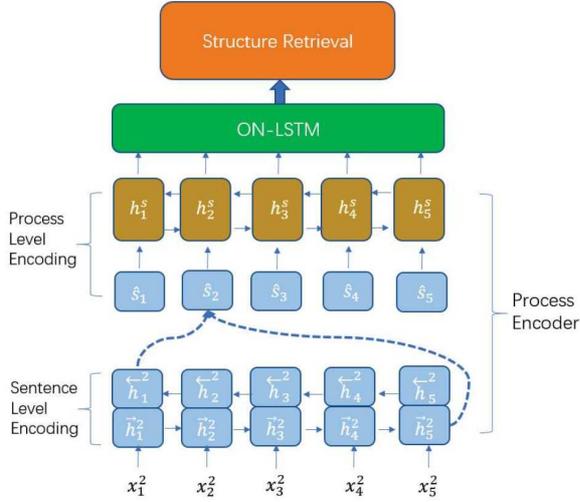

Figure 2: Architecture

**Document encoder**
We firstly introduce the process encoder component. We follow previous work (Qingyu Zhou and et al. 2018) in modeling documents hierarchically by first obtaining representations for sentences and then composing those into a process representation. In our approach, the process encoder consists of a sentence-level Recurrent Neural Network RNN-($RS$) and a document-level RNN-($RD$), which have the same structure. Given a paragraph of process description $PD = (S_1, S_2, ..., S_L)$ containing L sentences, each sentence $S_l$ in $PD$ is represented as $S_l = [w_{l1}, w_{l2}, ..., w_{ln}]$, where $w_{ln}$ means the $n_{th}$ word in $l_{th}$ sentence. $RS$ is applied to obtain a contextual representation for each word and get a basic sentence representation ŝ$_l$:

$$\hat{s}_l = [u_{l1}, u_{l2}, ..., u_{in}] = RS([w_{l1}, w_{l2}, ..., w_{ln}]) \quad (1)$$

Here we employ a bidirectional LSTM (Hochreiter and Schmidhuber 1997) as the recurrent unit, mathematically formulated as below:

$$i_t = \sigma(W^{ih}h_{t-1} + W^{ix}x_t + b^i) \quad (2)$$

$$f_t = \sigma(W^{fh}h_{t-1} + W^{fx}x_t + b^f) \quad (3)$$
$$o_t = \sigma(W^{oh}h_{t-1} + W^{ox}x_t + b^o) \quad (4)$$
$$g_t = \tanh(W^{gh}h_{t-1} + W^{gx}x_t + b^g) \quad (5)$$
$$c_t = f_t \odot c_{t-1} + i_t \odot g_t \quad (6)$$
$$h_t = o_t \odot \tanh(c_t) \quad (7)$$

where $x_t$ is the current input and h$_{t-1}$ is the hidden state of previous step. W$^{ih}$, W$^{fh}$, W$^{oh}$, W$^{gh}$ ∈ R$^{h×d}$, W$^{ix}$, W$^{fx}$, W$^{ox}$, W$^{gx}$ ∈ R$^{h×d}$, b$^i$, b$^f$, b$^o$, b$^g$ ∈ R$^h$ are the parameters to be learned. Notice that we follow the original definition of LSTM using t to represent the t$_{th}$ time-step, which corresponds to l$_{th}$ element in PD.

The biLSTM consists of a forward LSTM and a backward LSTM. For a sequence $x = x_1, ..., x_l$, a forward LSTM is used from left to right and gets a sequence of hidden states. The backward LSTM is used reversely, from right to left, and results in another sequence of hidden states.

$$\vec{h}_t = f(\vec{h}_{t-1}, x_t; \vec{\Theta}), t = 1, ..., l \quad (8)$$
$$\overleftarrow{h}_t = f(\overleftarrow{h}_{t-1}, x_t; \overleftarrow{\Theta}), t = 1, ..., l \quad (9)$$

Here $\vec{\Theta}$ and $\overleftarrow{\Theta}$ are parameters. We define $h_t = [\vec{h}_t, \overleftarrow{h}_t] \in R^{2h}$ which takes the concatenation of the hidden vectors from the RNNs in both directions. These representations can usefully encode information. After reading the words of the sentence $s_l$, we construct its sentence level representation $\hat{s}_l$ by concatenating the last forward and backward LSTM hidden vectors:

$$\hat{s}_l = [\vec{h}_l, \overleftarrow{h}_l] \quad (10)$$

We use another biLSTM as the process level encoder to read the sentences. With the sentence level encoded vectors ($\hat{s}_1$, $\hat{s}_2$, ..., $\hat{s}_l$) as inputs, the process level encoder does biLSTM encoding and produces two list of hidden vectors: ($\vec{s}_1, \vec{s}_2, ..., \vec{s}_L$) and ($\overleftarrow{s}_1, \overleftarrow{s}_2, ..., \overleftarrow{s}_L$). The process level representations $\tilde{s}_l$ of sentence $S_l$ is the concatenation of the forward and backward hidden vectors: $\tilde{s}_l = [\vec{s}_l, \overleftarrow{s}_l]$.

**Process-level language modeling with Ordered Neurons**
Recent findings in neural machine learning area show that the latent structure of a sentence can be captured by structural depths and distances (Hewitt and Manning 2019) which could be derived purely using the language modeling objective (Jie Hao and et al. 2019; Shen et al. 2019; Masaru Isonuma and Sakata 2019; Xing Wang and Shi 2019). ON-LSTM is one of the representative networks and is adopted in our approach. In this component, we train a process-level language model with ON-LSTM network. We firstly introduce the ON-LSTM network and then introduce the objective function.

**•ON-LSTM**
It is observed that natural language is hierarchically structured: smaller units (e.g., phrases) are nested within larger units (e.g., clauses). ON-LSTM proposes to add such structure-oriented inductive bias by ordering the neurons, which enables LSTM models to perform tree-like composition without breaking their sequential form (Shen et al. 2019). Ordered neurons enable dynamic allocation of neurons to represent different time-scale dependencies by controlling the update frequency of neurons. Compared with standard LSTM architecture, ON-LSTM introduces novel ordered neuron rules to update cell state, defined as below:

$$w_t = \tilde{f}_t \circ \tilde{\iota}_t \quad (11)$$
$$\hat{f}_t = f_t \circ w_t + (\tilde{f}_t - w_t) \quad (12)$$
$$\hat{\iota}_t = i_t \circ w_t + (\tilde{f}_t - w_t) \quad (13)$$
$$c_t = \hat{f}_t \circ c_{t-1} + \hat{\iota}_t \circ \hat{c}_t \quad (14)$$

where input gate $i_t$, forget gate ft and state $\hat{c}_t$ are same as that in the standard LSTM defined by Equation (2), (3) and (6) separately. The master forget gate $\hat{f}_t$ and the master input gate $\tilde{\iota}_t$ are newly introduced to ensure that when a given neuron is updated, all the neurons that follow it in the ordering are also updated. The product of the two master gates wt represents the overlap of $\hat{f}_t$ and $\tilde{\iota}_t$. Whenever the overlap exists (∃k, $w_{tk}$ > 0), the corresponding segment of neurons encodes are further controlled by the standard gates $f_t$ and $i_t$. (Shen et al. 2019) further introduced a new activation function CUMSUM as Equation (15) to find the splitting point $d$.

$$CU(\cdot) = CU\,M\,SU\,M(softmax(\cdot)) \quad (15)$$

Based on this activation function, the master gates are defined as Equation (16) and (17):

$$\tilde{f}_t = CU_f(x_t, h_{t-1}) \quad (16)$$
$$\tilde{\iota}_t = 1 - CU_f(x_t, h_{t-1}) \quad (17)$$

where $CU_f$ and $CU_i$ are two individual activation functions with their own trainable parameters.

• **Process-level language model objective function**
We adopt the document-level language model as (Dongyeop Kang and Hovy 2019) to represent the process structure. Adjacent sentences are treated as pairs for learning the standard seq2seq model. The objective is to maximize the likelihood of the current sentence given the previous sentence. The objective function is defined as Equation (18).

$$L = \sum_n log P(w_{ln}|w_{l,<n}, S_{l-1}) \quad (18)$$

**Structure Retrieve**
As introduced previously, the latent tree structure of process could be inferred from the trained ON-LSTM language model. Given a process description $PD = s_1, s_2, ..., s_l$, as input, each layer of the trained ON-LSTM language model infers a value $d_l$ for each $l = 1 ... L$ that measures the "level distance" between $s_{l-1}$ and $s_l$. An estimate of $d_l$ is calculated in Equation (19):

$$\hat{d}_l = E(d_l) = D_m - \sum_{k=1}^{D_m} \tilde{f}_{lk} \quad (19)$$

where $D_m$ is the size of the hidden state. $\tilde{f}_{lk}$ refers to the kth element in vector $\tilde{f}_l$ which is as Equation (16) (Refer to (Shen et al. 2019) for details). With this "level distance" sequences, greedy top-down retrieval algorithm (Mitchell Stern and Klein 2017) could be used to retrieve the latent tree structure.

# 4. Experiments
**Data**
We have a collection of nearly 100 PDD documents obtained from RPA project practice covering 52 target system applications (e.g. Outlook, SAP, Oracle and etc.). The customer sensitive information and noise texts (e.g. equations) are removed. Table 1 gives an overview of the characteristics of the resulting data set. The data from Table 1 illustrates that the average length of words per sentence is around 8. The longest process description sentence contains a total of 34 words. To avoid sparse, we abbreviated or split the sentences which have more than 15 words after the removal of punctuation and null elements while without changing its semantic. PDD documents in practical projects usually describe complex processes and therefore have deep structure. To make the training data more general and enlarge the total number of processes, we split the deep processes which have more than 6 layers. After splits, the total number of processes in the data set is around 2k. Average number of sentences per process is 12.

Table 1: Characteristics of PDD data set

| | |
|---|---|
| Number of the Documents | 100 |
| Average Number of words per sentence | 8.7 |
| Maximum Number of words per sentence | 34 |
| Average Number of Sentences per process | 12 |
| Number of Processes | 2k |

**Experimental Setup**
For the process-encoder, it is found that stacking BiLSTMs works better than a one-layer BiLSTM (Chen 2018). Therefore, we used 3 layers for the Stacked BiLSTMs to encode both the sentence-level and process-level inputs. For each input word, we firstly used word embedding to represent it as low-dimensional, real-valued vector. Since our training data set is relatively small, we initialized word embedding using GloVe word vectors (Jeffrey Pennington and Manning 2014), with the word embedding dimension 300, instead of fine tuning all the word embedding with BERT. For the ON-LSTM component, we used three-layer ONLSTM model2. We only modified the objective function which is defined as Equation (18). We followed the same hyper-parameters settings e.g., the embedding size, drop out parameters, as in the original ON-LSTM (Shen et al. 2019). The PDD data set is shuffled and then split as 90:10 into training and test sets.

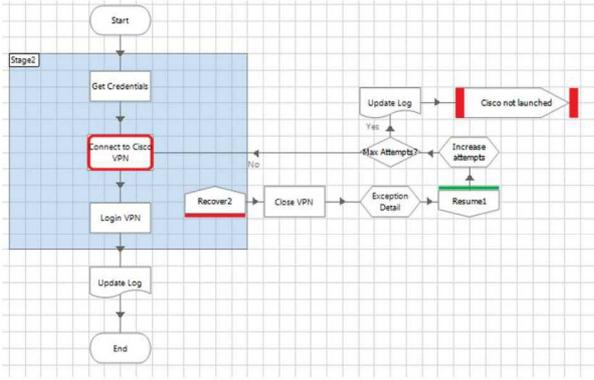 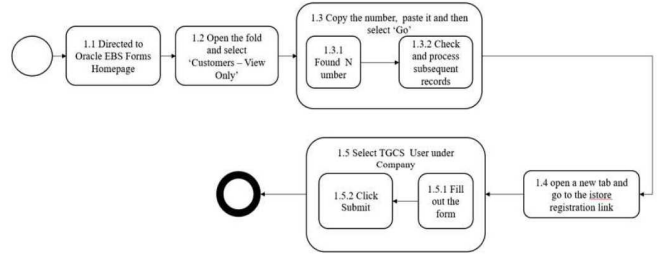

Figure 3: (a) BluePrism work flow diagram ;(b) BPMN example

**Results**

For the evaluation of process structure retrieval task, we need to compare the induced latent tree structure with the ground truth which is developed by human experts. The ground truth in our experiments were collected from two different resources separately: 1) Blueprism work flow. Part of the PDDs have corresponding RPA design flow developed using the Blueprism platform. 2) BPMN model. Some authors of PDD tend to describe processes in BPMN notation. Refer to Figure 3 for the example for each kind of graphical representations. We take advantage of BPMN model as ground truth during validation. If no BPMN model exists, we use Blueprism work flow instead.

Table 2: Result of the application of the evaluation metrics

| | |
|---|---|
| Edges | 57% |
| Nodes | 76% |
| simedg | 32% |

The evaluation results were based on the similarity between the ground truth and the retrieved tree structures. We employed a metric similar to the Graph Edit Distance in (Fabian Friedrich and Puhlmann 2011), defined as below:

$$simged(G_1, G_2) = 1 - (w_1 * sim_M + w_2 * sim_N + w_3 sim_E) \quad (20)$$

where,

$$sim_M = \frac{m^*}{|M|} \quad (21)$$

$$m^* = \begin{cases} \sum_{i=1}^{|M|} 1 - sim_{con}(Node_i), & if\ |M| > 0 \\ 1.0, & otherwise \end{cases} \quad (22)$$

$$sim_N = \frac{|\bar{N}_1| + |\bar{N}_2|}{|N_1| + |N_2|} \quad (23)$$

$$sim_E = \frac{|\bar{E}_1| + |\bar{E}_2|}{|E_1| + |E_2|} \quad (24)$$

In the above equations, $N_i$ is the set of nodes in graph $i$. $E_i$ is the set of edges in graph $i$. $|\bar{N}_i|$ is the set of nodes that have not been mapped to the other graph $i$. $|\bar{E}_i|$ is the not mapped set of edges in graph $i$. $i \in \{1, 2\}$ is the index for the graph pairs compared. $M$ represents the mapping of the nodes between the ground truth and the generated graph. $sim_{con}(Node_i)$ is the contextual similarity as defined in (R. Dijkman and et al. 2010). $w_1, w_2$ and $w_3$ are weights for the importance of the mapping, the unmapped nodes and the unmapped edges. For our experiments we gave the difference of mappings and edges slightly higher importance and assigned $w_1 = w_2 = w_3 = 0.3$ and $w_3 = 0.4$. To be noticed that, we treated one sentence as a node in the results of retrieved tree. The results are shown in Table 2.

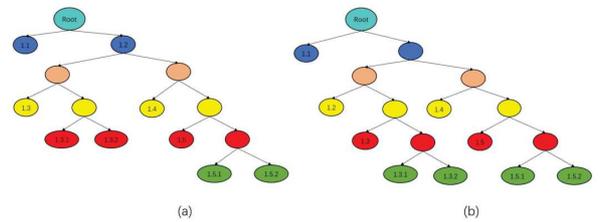

Figure 4: (a) CNN + PDD; (b) PDD

Rows Edges and Nodes measures the percentage of edges and nodes that are matched between the ground truth and retrieved. Simged measures the average percentage of similarity for the graph pairs. The results of un-matched nodes, are mainly caused by more detailed description in textual description, which can be partially explained by noise. During calculation of simged, we filtered such kind of nodes. The results show that our approach is able to correctly recreate 32% of the model in average. As the process documents data set is small, we also used CNN/Daily Mail data set (Jiatao Gu and Li 2016) which contains online news articles combined with the process data set to train the model. However, the performance didn't improve obviously. The retrieved tree structures for process illustrated in Figure 3(b) are as Figure 4. Figure 4(a) is based on the model with CNN mixed PDD data set as training data. Figure 4(b) is based on the model with PDD data set as training data.

## 5. Conclusion and Future Work

In this paper, we proposed an approach to retrieve the latent tree structure present in the textual business process documents from a trained ON-LSTM model, which uses the process-level language modeling objective. We evaluated the performance on data set of PDD, which is collected from the practical RPA projects. The results showed that our proposed approach could retrieve on average 32% of the process structure correctly measured by graph edit distance. Preliminary evaluation results show that it is promising to retrieve the latent hierarchical structure without any further human labeled knowledge. In organizations, there are large number of process description documents, which could be made use of as training data to further improve the performance.

However, there are still many spaces to improve the proposed approach to reach better performance in addition to adding training data. Firstly, for the input representation, we just used word embedding to represent the word in process sentences. There are many other semantic or syntactic features which reflect the properties of word in its context, such as its part-of-speech (POS), named entity recognition (NER) tags could be added. Secondly, we used two-level LSTM based process encoder to encode the process. Transformer model proposed by Google researchers (Ashish Vaswani and et al. 2017) proved to be efficient in encoding documents (Yang Liu and Lapata 2019). Finally, restricted by the ONLSTM based model, the retrieve algorithm could only generate binary tree, which can't reflect the complex process flow structure. Future work should be focused on the augmentation of the ON-LSTM model to make it suitable to learn the process structure.